\documentclass[10pt,twocolumn,letterpaper]{article}

\usepackage{cvpr}              
\definecolor{cvprblue}{rgb}{0.21,0.49,0.74}
\usepackage[pagebackref,breaklinks,colorlinks,allcolors=cvprblue]{hyperref}
\usepackage{amsmath} 
\def\paperID{} 
\def\confName{CVPR}
\def\confYear{2026}

\title{IMU-HOI: A Symbiotic Framework for Coherent Human-Object Interaction and Motion Capture via Contact-Conscious Inertial Fusion
}

\author{Lizhou Lin$^1$\thanks{Equal contribution}
\quad
Songpengcheng Xia $^{1*}$
\quad
Zengyuan Lai $^1$ 
\quad
Lan Sun$^1$
\quad
Jiarui Yang$^1$
\quad
Ling Pei$^{1}$ \thanks{Corresponding authors \\ This work was supported by the National Nature Science Foundation of China (NSFC) under Grant 62273229.} 
\\
$^1$Shanghai Jiao Tong University
}

\usepackage{booktabs,multirow}
\usepackage{array} 
\usepackage{cuted} 
\usepackage{graphicx}
\usepackage{subcaption}
\begin{document}

\maketitle
\begin{abstract}

Capturing full-body human motion with object interactions is crucial for AR/VR and robotics applications, yet it remains challenging for conventional vision-based methods due to occlusions and constrained capture volumes. Inertial measurement units (IMUs) offer a compelling alternative without line-of-sight requirements, but existing IMU-based motion capture assumes an isolated human and ignores object contacts and dynamics. To bridge this gap, we present IMU-HOI, a novel framework that jointly recovers full-body human pose and 6-DoF object trajectory from sparse IMUs on the body and object, explicitly modeling human-object Interaction.
Our approach first infers probabilistic hand–object contacts directly from IMU streams and uses them as a high-level signal to route between kinematic and inertial reasoning. These contact cues drive a three-stage fusion pipeline that refines human pose and root translation, and fuses hand-based forward kinematics with object-IMU integration for object motion, yielding coherent, drift-resilient trajectories for both human and object.  Experiments on challenging human-object interaction scenarios demonstrate substantial accuracy gains over prior inertial motion capture methods. Moreover, IMU-HOI can be plugged into existing sparse-IMU mocap backbones with minimal changes, effectively extending the scope of purely inertial motion capture from isolated humans to full human–object interaction and joint motion estimation.

\end{abstract}

\section{Introduction}
\label{sec:intro}
The capture of full-body human motion \emph{together with} object interactions is a cornerstone for advancing embodied AI applications, such as immersive AR/VR and seamless human–robot collaboration~\cite{wang2025ego4o, xia2025envposer, dai2024hmd, Guzov_2021_CVPR, lin2025let, xu2025intermimic, xu2025interact}. While vision-based motion capture has been the predominant approach, its efficacy diminishes in natural environments due to persistent challenges such as occlusions, limited capture volume, and viewpoint dependence, which are exacerbated during dynamic motions and object manipulations~\cite{Gilbert2019IJCV, Zhang2020CVPR, Kaichi2020Sensors}. 
Inertial Measurement Units (IMUs) offer a compelling alternative: they enable 3D human pose estimation (HPE) from a sparse set of wearable sensors, unshackled from line-of-sight and lighting constraints~\cite{mollyn2023imuposer, xu2024mobileposer, Ren2023LidarPoser, Armani2024UltraPoser, Xue2025GIP, ilic2025garment, lee2024mocapevery}. Pioneering systems (e.g., DIP~\cite{huang2018deep}, TransPose~\cite{yi2021transpose}, TIP~\cite{jiang2022transformer}) have demonstrated remarkable performance by leveraging learned kinematic priors and simplistic physical constraints (e.g., foot-ground contact) to mitigate inherent sensor noise and drift.

\begin{figure}[t]
  \centering
  \includegraphics[width=\linewidth]{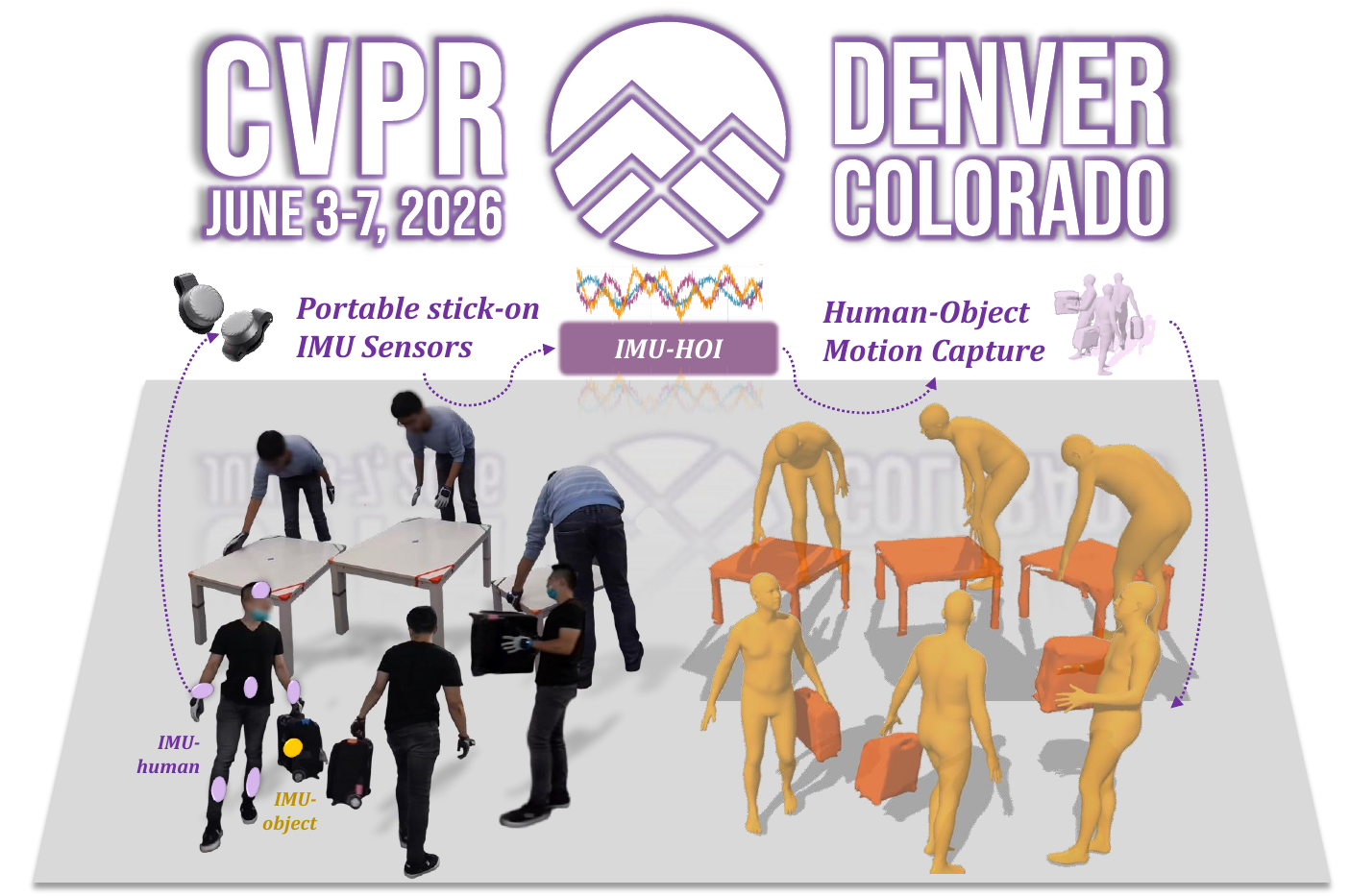}
  \vspace{-1.5em}
  \caption{\textbf{IMU-HOI} recovers full-body human motion and 6-DoF object trajectories from a sparse set of stick-on IMUs attached to both the body and the object, without using cameras.}
  \label{intro_fig}
  \vspace{-1.5em}
\end{figure}

However, a fundamental limitation persists: existing inertial motion capture methods operate under the assumption of an \emph{isolated} human body~\cite{huang2018deep, zhang2024dynamic}. They ignore the rich context of human–object interactions (HOI), neither modeling contact nor estimating the object’s state. This overlooks a critical dimension of real-world activities, where interactions with tools, sports equipment, and everyday objects are ubiquitous~\cite{bhatnagar2022behave, zhang2024hoim3, zhao2024imhoi, liu2022hoi4d, lu2025humoto, guzov2024interactionreplica}. Consequently, there remains a significant gap between the output of these systems and the true dynamics of embodied interaction.

Intriguingly, the form factor of IMUs makes them uniquely suited to bridge this gap. Just as they are worn on the body, IMUs can be flexibly attached to everyday objects, enabling instrumented HOI in settings where cameras are impractical or undesirable, as shown in Fig.~\ref{intro_fig}. This opens the door to robust, in-the-wild motion capture of interactive scenes using only sparse inertial sensors. Yet, this direction remains largely unexplored: existing IMU-based methods disregard manipulated objects altogether, while vision-based HOI capture typically relies on complex multi-view rigs, pre-scanned meshes, or carefully controlled environments~\cite{bhatnagar2022behave, zhang2024hoim3, zhao2024imhoi, liu2022hoi4d, lu2025humoto, guzov2024interactionreplica}


To address this critical gap, we propose \textbf{IMU-HOI}, a symbiotic framework that jointly infers full-body human pose and 6-DoF object trajectory from sparse IMUs affixed to both the body and the object, with explicit reasoning about human-object contacts. Concretely, our pipeline proceeds in three stages. First, a learning-based module estimates short-horizon motion cues and probabilistic contact events from body- and object-IMUs, identifying when and which hand is in contact and producing calibrated contact priors. Second, we refine the human pose and root translation using a multi-path inertial fusion module: building on a state-of-the-art sparse-IMU poser, we adopt a part-based design and a two-branch root-translation head that exploits both foot contact and trunk dynamics, yielding stable full-body joints and wrist trajectories. Third, we estimate the object trajectory via a contact-conditioned, three-branch fusion: two kinematic branches reconstruct object motion from hand forward kinematics (FK) under a quasi-rigid contact model, while an IMU branch integrates object accelerations with a residual network to correct for drift. A contact-gated fusion module combines these hypotheses in a Bayesian-style manner, anchoring the object to the hand when contact is confident and falling back to IMU when it is not. This closed-loop design yields coherent, physically plausible motion capture for both human and object.


In summary, our main contributions are threefold:

\begin{itemize}

 \item We introduce, to our knowledge, the first purely inertial framework for coherent human–object motion capture: IMU-HOI formulates HOI tracking from sparse body- and object-IMUs as a unified problem, and treats hand–object contact as a first-class probabilistic signal that routes between kinematic and inertial reasoning throughout the pipeline.
    
\item We develop a three-stage, contact-conscious inertial fusion architecture that seamlessly integrates object interaction cues into both human pose estimation and object translation, combining a part-based human module with a contact-gated FK–IMU fusion module for robust, drift-resistant HOI tracking.

\item Extensive experiments on three challenging HOI benchmarks demonstrate that IMU-HOI consistently improves object translation and interaction fidelity over strong IMU-based baselines. Moreover, we show that our contact-gated fusion is modular and plug-and-play: when instantiated with existing sparse-IMU pose estimators, it systematically upgrades them with interaction-aware object dynamics tracking at minimal additional cost.

\end{itemize}

\section{Related Work} \label{sec:relatedwork}

\subsection{Human Pose Estimation with Wearables} 
Inertial motion capture (mocap) avoids issues like occlusion and lighting, evolving from early database-driven reconstruction with sparse accelerometers~\cite{slyper2008action, Tautges_2011} to SMPL-based optimization and real-time learning. Commercial systems like Xsens~\cite{schepers2018xsens} deliver high accuracy but require many sensors, limiting scalability. A key milestone was sparse IMU-to-SMPL fitting~\cite{von2017sparse}, followed by learning-based regression of SMPL parameters from IMUs~\cite{huang2018deep}. Hierarchical BiRNN pipelines with physics-aware refinement improved robustness~\cite{yi2021transpose, yi2022physical}, and stationary body point (SBP) reasoning reduced drift~\cite{jiang2022transformer}. Recent advances improve priors and robustness: DynaIP~\cite{zhang2024dynamic} learns part-aware motion and uses pseudo-velocity cues; Fast Inertial Poser~\cite{xiao2024fast} achieves real-time tracking with causal modeling and shape conditioning; PNP~\cite{yi2024physical} models fictitious forces to leverage accelerations; and recent works~\cite{mollyn2023imuposer, xu2024mobileposer, sun2025suite} reduce sensor count using commodity wearables. Recent state-space models~\cite{kim2025probabilistic, zhao2025kinest, zhao2025ssd} further improve tracking robustness and uncertainty estimation from sparse inputs. 
Despite these advances, all the above methods focus solely on reconstructing the human body: the manipulated objects, their states and contacts remain unobserved, and global alignment is only weakly constrained. In this work, we explicitly model the object side and human–object interaction, extending purely inertial motion capture from body-only tracking to coherent human–object motion capture.

\subsection{Multi-Modal HOI Datasets and Capture} 
3D HOI is widely studied under external-camera and simulation setups.
On the data side, multi-view datasets~\cite{bhatnagar2022behave,zhang2024hoim3,huang2024intercap}, simulation-based datasets~\cite{wang2025pahoi}, and hand--object grasp collections~\cite{taheri2020grab} provide high-fidelity supervision.
However, they typically rely on dense camera rigs, markers, or controlled studios, and often assume static scenes or pre-scanned CAD models~\cite{hassan2019prox}, limiting deployment in everyday environments with movable or unknown objects.

To move beyond studio capture, recent methods estimate HOI from monocular RGB in the wild.
PHOSA~\cite{zhang2020phosa} recovers static human--object configurations via contact and collision cues, while learning-based approaches~\cite{xie2022chore,xie2022vistracker} model dynamic interactions using motion priors, visibility-aware tracking, or generative diffusion~\cite{li2023omghm}. 
Yet camera-based solutions remain sensitive to occlusions, often require calibrated external hardware and known object geometry, and raise privacy and scalability concerns in real-world deployments~\cite{bhatnagar2022behave,zhang2024hoim3,zhang2020phosa,xie2022chore,xie2022vistracker,li2023omghm}.

These challenges have motivated sensor-assisted HOI capture that moves sensing onto the actor.
Egocentric HOI has grown with AR/VR, leading to first-person datasets~\cite{xu2025firstperson} and head-mounted capture methods: EgoChoir~\cite{yang2024egochoir} infers 3D contacts and affordances from head motion and egocentric video, and Lost\&Found~\cite{behrens2025lostfound} tracks scene-graph changes from wearable sensors.
In parallel, sparse IMU-based pose estimation~\cite{huang2018deep,jiang2022transformer} has advanced but largely ignores objects, while hybrid methods combine vision and inertial data: I'M~HOI~\cite{zhao2024imhoi} fuses an object IMU with monocular video, and HybridCap~\cite{liang2023hybridcap} uses body IMUs to stabilize monocular motion capture.
Most closely related to our setting, Interaction Replica~\cite{guzov2024interactionreplica} shows that full-body mocap suits can infer object trajectories from human motion alone, and ECHO~\cite{petrov2025echo} reconstructs body, object, and contact from head and wrist sensors with a diffusion transformer.

In contrast to these methods, which often rely on visual input, detailed object models, or extensive sensor suites, our work demonstrates that accurate full HOI pose can be recovered from a sparse set of only six body IMUs and a single object IMU, without any visual cues.

\begin{figure*}[t]
    \centering
    \includegraphics[width=1\linewidth]{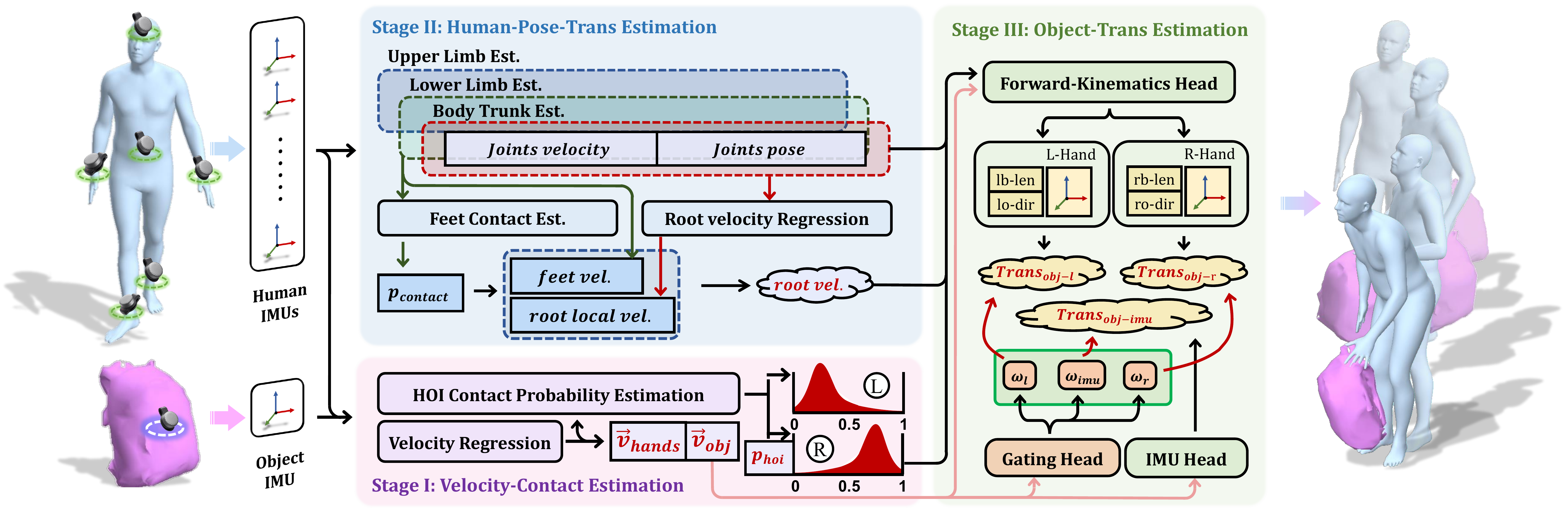}
    \vspace{-1.5em}
    \caption{Overview of our three-stage HOI pipeline from sparse human–object IMUs: Stage I (mid-bottom) predicts hand/foot contacts and object velocity, Stage II (mid-top) estimates part-based human pose and root translation, and Stage III (mid-right) contact-gates FK and IMU branches to recover object translation.}
    \label{fig:pipeline}
    \vspace{-1.5em} 
\end{figure*}

\section{Method}
\label{sec:method}

\subsection{Problem Formulation and Overview}
We address human--object interaction (HOI) reconstruction from sparse body and object IMUs with a three-stage architecture. Stage~I estimates a calibrated hand--object contact prior and short-horizon object velocity from local inertial cues; Stage~II predicts a temporally consistent full-body pose and root translation using a DynaIP-style part-based backbone with a TransPose-inspired, contact-aware translation head; Stage~III recovers global object translation by fusing two contact-conditioned FK branches with an IMU-integrated trajectory. Across all stages, contact is modeled as an explicit probabilistic signal that gates kinematic versus inertial hypotheses at each frame and is optimized in a staged curriculum, yielding an interpretable and drift-resistant HOI estimator.

At time $t$, we observe human IMUs $\mathbf{X}^{\text{hum}}_t \in \mathbb{R}^{N_{\text{hum}}\times 9}$ and an object IMU $\mathbf{X}^{\text{obj}}_t \in \mathbb{R}^{9}$, together with an initial state
\begin{equation}
\mathcal{S}_0 = \big\{\mathbf{pose}_0,\ \mathbf{p}_{\mathrm{root}}(0),\ {}^{W}\!R_O(0),\ \mathbf{p}_O(0)\big\},
\end{equation}
where ${}^{W}\!R_O(t)\in SO(3)$ denotes the object orientation in the world frame and $\mathbf{p}_{\mathrm{root}}(t),\mathbf{p}_O(t)\in\mathbb{R}^3$ denote the world-space root and object positions.

Our model outputs, for each frame, full-body joints $\hat{\mathbf{J}}_t\in\mathbb{R}^{24\times 3}$, a root position trajectory $\hat{\mathbf p}_{\mathrm{root}}(t)$, and an object position trajectory $\hat{\mathbf p}_O(t)$ obtained by fusing three hypotheses with contact-gated weights $w_{t,k}$:
\begin{equation}
\hat{\mathbf p}_O(t) = \sum_{k\in\{L,R,IMU\}} w_{t,k}\,\hat{\mathbf p}^{(k)}_O(t),
\end{equation}
where
$\hat{\mathbf p}^{FK\!-\!L}_O(t)$ and
$\hat{\mathbf p}^{FK\!-\!R}_O(t)$
denote the object positions predicted via forward kinematics from the left- and right-hand kinematic chains, respectively, and
$\hat{\mathbf p}^{IMU}_O(t)$
is the object position inferred from the object IMU branch.
Unless otherwise stated, all reported metrics are computed on
$\hat{\mathbf{J}}_t$,
$\hat{\mathbf p}_{\mathrm{root}}(t)$,
and $\hat{\mathbf p}_O(t)$.

\subsection{Stage~I: Contact and Velocity Estimation}
\label{subsec:stage1}

\medskip\noindent\textbf{Contact representation.}
We model the momentary contact state as a categorical distribution over
$\{L, R, 0\}$ (left hand, right hand, no contact). Instead of a standard
3-way softmax, we parameterize
\begin{equation}
\boldsymbol{\pi}_t
=
\big[p_L(t),\ p_R(t),\ 1-\max(p_L(t),p_R(t))\big] \in \Delta^2,
\end{equation}
where $p_L(t)$ and $p_R(t)$ denote the marginal probabilities of left-hand and right-hand contact at time $t$. It (i) preserves probability mass when neither hand is in contact, (ii) avoids double counting during transitions or bimanual touches by collapsing shared mass into the complement, and (iii) maps directly to the three object-translation sources used in Stage~III (left-FK, right-FK, IMU).

A compact recurrent head (LSTM) takes a short temporal window of human and object IMUs together with intermediate motion cues and outputs logits for $p_L(t)$ and $p_R(t)$. We train this head with focal cross-entropy~\cite{lin2017focal} to handle long no-contact stretches and short contact spikes, and calibrate $\boldsymbol{\pi}_t$ as a prior for fusion using temperature scaling plus a light regularization term (details in the supplement).

\medskip\noindent\textbf{Object velocity estimation.}
In parallel, a second head regresses the object linear velocity
$\hat{\mathbf v}_O(t)$ from the object IMU sequence and orientation increments $\Delta {}^{W}\!R_O(t)$. We use a robust regression loss (Huber)~\cite{huber1992robust} together with a short-horizon integral consistency term that ties predicted velocity to observed displacement over small windows, improving temporal smoothness without sacrificing responsiveness. A differentiable object-stationarity gate further down-weights both $\|\hat{\mathbf v}_O(t)\|$ and spurious contact logits when the object IMU indicates near-zero motion, reducing false positives in quiescent segments.

Stage~I thus provides two key signals for downstream reasoning:
(i) a calibrated contact prior $\boldsymbol{\pi}_t$, and
(ii) a drift-aware, short-horizon object velocity $\hat{\mathbf v}_O(t)$.

\subsection{Stage~II: Part-Based Human Pose and Root Translation Estimation}
\label{subsec:stage2}

Stage~II estimates a temporally consistent human pose and root translation from human IMUs. As illustrated in Fig.~\ref{fig:pipeline} (mid-top), the module is implemented as a unified Human-Pose-Module that combines a part-based pose backbone with a two-branch translation head.

\medskip\noindent\textbf{Part-based pose backbone.}
Following DynaIP~\cite{zhang2024dynamic}, we decompose the body into three regions (lower limbs, trunk, upper limbs) and partition the IMUs accordingly. A lightweight global RNN first summarizes the sequence $\mathbf{X}^{\text{hum}}_{1:T}$ into a per-frame context feature. For each region, a dedicated \emph{SubPoser} then takes the region-specific IMUs and global context as input and outputs (i) local joint velocities and (ii) a reduced set of global joint rotations for that region. Concatenating the three SubPosers yields a reduced global pose.
To obtain a full 24-joint body, we insert the reduced rotations into a standard SMPL body model, complement them with calibrated sensor orientations for the remaining bones, and apply differentiable forward kinematics (FK). This produces world-space joint positions $\hat{\mathbf{J}}_t$, which serve as both the final pose output of Stage~II and an intermediate representation for root translation and wrist trajectories $\hat{\mathbf p}^{(L)}_H(t),\hat{\mathbf p}^{(R)}_H(t)$.

\medskip\noindent\textbf{Two-branch root translation.}
For root translation, we follow the spirit of TransPose~\cite{yi2021transpose}, which estimates global motion from six IMUs via two complementary branches: a contact-driven branch and an IMU-driven branch. Building on this design, we introduce targeted modifications that further improve the accuracy of root translation estimation.

\medskip\noindent\textbf{Lower-limb contact branch.}
The Lower-limb contact branch reuses the lower-limb velocities predicted by the corresponding SubPoser. We extract the velocities of both feet and the root, and feed them, together with foot IMUs and global context, into an RNN that outputs foot-contact probabilities. Following TransPose, we derive a contact-driven root-velocity hypothesis by selecting the more likely contacting foot and mapping its FK velocity to root motion, with a small learned bias to account for residual modeling errors.

\medskip\noindent\textbf{Trunk-velocity branch.}
Instead of feeding all joint positions to Local-root-velocity branch as in TransPose, we construct a compact \emph{trunk-velocity descriptor} from the DynaIP outputs by selecting trunk joints (hips, spine, neck, head) and computing finite-difference velocities. Trunk-velocity branch takes this descriptor, together with IMUs around the root/head and the global context, and predicts a local root velocity that is rotated into world coordinates using the root IMU orientation. Using velocities rather than positions provides a cleaner signal and is closer to the prediction target.

\medskip\noindent\textbf{Contact-aware fusion.}
Given the world-space root velocities from the two branch, we compute a scalar contact confidence from the maximum foot-contact probability and convert it into a fusion weight $w_t\in[0,1]$. The two hypotheses are blended in a convex combination and integrated over time with floor constraints to prevent penetration. Adding this translation back to FK joint positions yields world-space joint trajectories and, in particular, stable wrist positions that are passed to Stage~III.

Stage~II uses standard losses from DynaIP and TransPose for supervision, including joint rotation/position, velocity of each part, foot contact probability, and root velocity/translation, which we will describe in detail in the supplementary material.

\subsection{Stage~III: Object Translation via Contact-Gated FK--IMU Fusion}
\label{subsec:stage3}

Stage~III estimates the global object translation by fusing two kinematic (FK) branches, conditioned on the left/right hands, with an IMU-driven branch. The design is built around a \emph{quasi-rigid contact} assumption.

\medskip\noindent\textbf{Quasi-rigid contact model.}
Once a hand $s\in\{L,R\}$ establishes contact with the object, we assume that the corresponding contact point on the object remains fixed in the object frame over the interaction segment. Let ${}^W\!\mathbf{p}^{(s)}_H(t)$ be the world-space position of hand $s$ and ${}^W\!\mathbf{p}_O(t)$ the object IMU position. Then
\begin{equation}
{}^O\!\mathbf{d}^{(s)} = {}^W\!R_O(t)^\top\big({}^W\!\mathbf{p}_O(t) - {}^W\!\mathbf{p}^{(s)}_H(t)\big),
\end{equation}
is constant over time, and
\begin{equation}
{}^W\!\mathbf{p}_O(t) = {}^W\!\mathbf{p}^{(s)}_H(t) + {}^W\!R_O(t)\,{}^O\!\mathbf{d}^{(s)}.
\end{equation}

In our setting, no object mesh is available; hence we cannot refine the contact point geometrically and instead learn an object-frame offset that implicitly encodes the contact configuration.

\medskip\noindent\textbf{Kinematic FK branches.}
For each hand $s\in{L,R}$, a lightweight recurrent head takes as input the hand kinematics and object orientation, and predicts an object-frame unit direction ${}^O\hat{\mathbf u}^{(s)}_t \in \mathbb{S}^2$ and a non-negative reach length $\hat{\ell}^{(s)}_t \ge 0$. We interpret
${}^{O}\hat{\mathbf u}^{(s)}_t\,\hat{\ell}^{(s)}_t$
as an approximation of the constant offset ${}^O\!\mathbf{d}^{(s)}$, and obtain a kinematic object-translation hypothesis
\begin{equation}
\hat{\mathbf p}^{FK\!-\!s}_O(t)
=
\hat{\mathbf p}^{(s)}_H(t)
+
{}^{W}\!R_O(t)\,{}^{O}\hat{\mathbf u}^{(s)}_t\,\hat{\ell}^{(s)}_t.
\end{equation}

Predicting offsets in the object frame factors out global rotation and stabilizes learning across viewpoints and grasp orientations. During training, we supervise these offsets whenever hand--object contact is annotated, using separate length/direction losses and a hand--object offset loss in the object frame.

\medskip\noindent\textbf{IMU branch with residual integration.}
In parallel, given the object velocity sequence $\hat{\mathbf v}_O(t)$ from Stage-I, we construct an IMU-integrated trajectory $\tilde{\mathbf p}^{IMU}_O(t)$ by applying a residual update to the previous fused position:
\begin{equation}
\tilde{\mathbf p}^{IMU}_O(t) =
\hat{\mathbf p}_O(t-1) + \hat{\mathbf v}_O(t)\,\Delta t,
\end{equation}
where $\Delta t = 1/\text{fps}$. This branch is supervised by an object translation loss.

\medskip\noindent\textbf{Contact-gated fusion and consistency.}
The final object trajectory is obtained via contact-aware fusion of the left/right FK branches and the IMU branch. A causal LSTM observes object IMU features and outputs fusion logits $\mathbf z_t \in \mathbb{R}^3$. The Stage-I contact prior $\boldsymbol{\pi}_t$ is injected as a multiplicative bias:
\begin{equation}
\mathbf w_t = \mathrm{softmax}\!\Big(\tfrac{1}{\tau}\big[\mathbf z_t + \beta\,\log \boldsymbol{\pi}_t\big]\Big),
\end{equation}
where $\tau>0$ is a temperature hyperparameter controlling the sharpness of the gating, and $\beta$ balances the influence of the contact prior $\boldsymbol{\pi}_t$, and the fused position is
\begin{equation}
\hat{\mathbf p}_O(t)
=
\sum_{k\in\{L,R,IMU\}} w_{t,k}\,\hat{\mathbf p}^{(k)}_O(t).
\end{equation}

This Bayesian-style routing yields interpretable behavior: confident left-hand contact amplifies the left-FK branch, while no-contact naturally shifts weight to the IMU branch. We regularize rapid changes in $\mathbf w_t$ and lightly smooth FK outputs to stabilize fusion during contact transitions, without over-smoothing sharp corrections.

Finally, we enforce kinematic--inertial consistency between the fused positions $\hat{\mathbf p}_O(t)$, Stage-I velocities $\hat{\mathbf v}_O(t)$, and IMU accelerations via simple finite-difference losses on first- and second-order differences, which suppress jitter and constrain drift.

\subsection{Optimization and training schedule}
We adopt a staged curriculum and freeze earlier modules once trained. \textbf{Stage~I} is first trained in isolation with
\begin{equation}
\mathcal{L}^{(1)} = \mathcal{L}_{\text{hands}} + \lambda_{\text{vel}}\mathcal{L}_{\text{vel}} + \lambda_{\text{cal}}\mathcal{L}_{\text{cal}},
\end{equation}
where $\mathcal{L}_{\text{hands}}$ and $\mathcal{L}_{\text{vel}}$ are the focal-contact and velocity losses defined above, and $\mathcal{L}_{\text{cal}}$ denotes the light calibration regularization (temperature scaling, KL prior and the object-stationarity gate). After convergence, Stage~I is frozen for all subsequent training.

\textbf{Stage~II} is then trained on top of the frozen Stage~I. Its loss could be represented by
\begin{equation}
\mathcal{L}^{(2)} = \mathcal{L}_{\text{pose}} 
+ \lambda_{\text{root}}\mathcal{L}_{\text{root}} 
+ \lambda_{\text{part}}\mathcal{L}_{\text{vel-part}} 
+ \lambda_{\text{feet}}\mathcal{L}_{\text{feet}}.
\end{equation}

It combines standard supervision on joint rotations/positions ($\mathcal{L}_{\text{pose}}$), root velocity/translation ($\mathcal{L}_{\text{root}}$), per-part velocities ($\mathcal{L}_{\text{vel-part}}$) and feet-contact probabilities ($\mathcal{L}_{\text{feet}}$). This yields stable human pose, root trajectory and wrist positions that are kept fixed when training Stage~III (unless explicitly stated otherwise).

\newcommand{\cw}{0.88cm}                 
\newcommand{\mw}{1.5cm}                  
\newcolumntype{C}{>{\centering\arraybackslash}m{\cw}}
\newcolumntype{L}{>{\raggedright\arraybackslash}m{\mw}}
\begin{table*}[t]
\centering
\caption{Comparison on three HOI datasets. Numbers are lower-is-better. Best in \textbf{bold}.}
\vspace{-3mm}
\label{tab:sota_without_root}
\setlength{\tabcolsep}{2pt}
\renewcommand{\arraystretch}{1.06}
\scriptsize
\resizebox{\linewidth}{!}{%
\begin{tabular}{L*{15}{C}}
\toprule
& \multicolumn{5}{c}{OMOMO} 
& \multicolumn{5}{c}{$IMHD^{2}$} 
& \multicolumn{5}{c}{BEHAVE} \\
\cmidrule(lr){2-6}\cmidrule(lr){7-11}\cmidrule(lr){12-16}
\textbf{Methods} 
& Obj Err  & HOI Err  & Ang Err  & Pos Err  & Jitter
& Obj Err  & HOI Err  & Ang Err  & Pos Err  & Jitter
& Obj Err  & HOI Err  & Ang Err  & Pos Err  & Jitter \\
\midrule
DIP* 
& 25.84 & 26.79 & 3.05 & 3.34 & 2.74
& 90.37 & 93.09 & 7.36 & 8.97 & 9.57
& 38.91 & 39.31 & 4.79 & 4.70 & 7.04 \\
TIP* 
& 44.41 & 48.01 & 5.48 & 8.55 & 5.77
& 92.75 & 106.74 & 10.84 & 49.23 & 66.88
& 26.43 & 30.24 & 7.86 & 11.97 & 8.47 \\
TransPose* 
& 32.54 & 32.73 & 4.48 & 3.15 & \textbf{2.47}
& 90.37 & 90.61 & 7.76 & 5.27 & 9.57
& 40.39 & 40.40 & 6.71 & 3.71 & 8.58 \\
GlobalPose* 
& 39.34 & 39.51 & 4.13 & 3.69 & 4.04
& 101.27 & 102.03 & 6.61 & 8.31 & \textbf{8.17}
& 40.14 & 40.21 & 6.00 & 4.73 & \textbf{6.88} \\
\textbf{Ours (Fusion)} 
& \textbf{14.15} & \textbf{14.94} & \textbf{2.84} & \textbf{2.27} & 2.75
& \textbf{49.76} & \textbf{51.09} & \textbf{4.51} & \textbf{3.85} & 10.57
& \textbf{22.26} & \textbf{22.62} & \textbf{4.52} & \textbf{2.72} & 8.20 \\
\bottomrule
\end{tabular}%
}
\vspace{-0.6em}
\end{table*}

\renewcommand{\cw}{0.93cm}   
\renewcommand{\mw}{1.5cm}   
\begin{table*}[t]
\centering
\caption{Main comparison with root-translation and jitter. Numbers are lower-is-better. Best in \textbf{bold}.}
\vspace{-3mm}
\label{tab:sota_with_root}
\setlength{\tabcolsep}{2pt}
\renewcommand{\arraystretch}{1.05}
\scriptsize
\resizebox{\linewidth}{!}{%
\begin{tabular}{L*{18}{C}}
\toprule
& \multicolumn{5}{c}{OMOMO}
& \multicolumn{5}{c}{$IMHD^{2}$}
& \multicolumn{5}{c}{BEHAVE} \\
\cmidrule(lr){2-6}\cmidrule(lr){7-11}\cmidrule(lr){12-16}
\textbf{Methods}
& Obj Err & HOI Err & Pos Err & Trans Err & Jitter 
& Obj Err & HOI Err & Pos Err & Trans Err & Jitter 
& Obj Err & HOI Err & Pos Err & Trans Err & Jitter  \\
\midrule
TransPose*
& 32.09 & 40.43 & 6.36 & 19.36 & 6.09
& 94.57 & 100.65 & 5.87 & 37.79 & 10.66
& 50.70 & 55.83 & 6.05 & 23.52 & 11.76 \\
GlobalPose*
& 39.31 & 40.35 & 3.67 & \textbf{10.80} & 3.94
& 103.27 & 108.33 & 8.29 & 58.36 & \textbf{8.15}
& 42.38 & 45.47 & 4.71 & 16.05 & \textbf{6.63} \\
\textbf{Ours (Fusion)}
& \textbf{11.31} & \textbf{17.44} & \textbf{2.35} & 11.07 & \textbf{2.77}
& \textbf{43.97} & \textbf{43.39} & \textbf{3.69} & \textbf{18.37} & 10.38
& \textbf{20.90} & \textbf{25.74} & \textbf{2.77} & \textbf{11.14} & 8.19 \\
\bottomrule
\end{tabular}%
}
\vspace{-0.5em}
\end{table*}

\textbf{Stage~III} is optimized in two phases. In the first phase we freeze both Stage~I and Stage~II and train only the object-translation module with
\begin{equation}
\mathcal{L}^{(3)} = \mathcal{L}_{\text{trans}} 
+ \lambda_{\text{cons}}\mathcal{L}_{\text{cons}}
+ \lambda_{\text{HOI}}\mathcal{L}_{\text{HOI}}
+ \lambda_{\text{smooth}}\mathcal{L}_{\text{smooth}},
\end{equation}
where $\mathcal{L}_{\text{trans}}$ is the object position regression (obj\_trans), $\mathcal{L}_{\text{cons}}$ collects the velocity/acceleration consistency terms (obj\_vel\_cons, obj\_acc\_cons), $\mathcal{L}_{\text{HOI}}$ aggregates the length and direction losses on the hand–object offsets for both hands as well as an HOI error term that directly supervises the hand–object relative pose, and $\mathcal{L}_{\text{smooth}}$ regularizes the temporal variation of fusion weights and FK outputs. After Stage~III has converged, we \emph{unfreeze Stage~II but keep Stage~I frozen}, and perform a short joint fine-tuning of Stage~II+III with a reduced learning rate and total loss
\begin{equation}
\mathcal{L}^{(2+3)} = \mathcal{L}^{(2)} + \lambda_{\text{joint}}\mathcal{L}^{(3)},
\end{equation}
which slightly adjusts human pose and root translation to better support object translation, while preserving the calibrated contact priors from Stage~I. In all experiments, Stage~I is only updated in its dedicated first stage and remains frozen thereafter.






\section{Experiment}
\label{sec:experiments}
We organize our experimental analysis as follows:
(1) \emph{Experimental setup}: datasets, metrics, and implementation details;
(2) \emph{Comparison with state-of-the-art}: quantitative and qualitative benchmarking;
(3) \emph{Ablation studies}: analysis of each proposed component.

\subsection{Experimental Setup}




\medskip\noindent\textbf{Datasets.}
We evaluate on three public HOI benchmarks: OMOMO~\cite{li2023omghm}, BEHAVE~\cite{bhatnagar2022behave}, and IMHD$^{2}$~\cite{zhao2024imhoi}. 
To match temporal statistics, IMHD$^{2}$ is downsampled from 60\,fps to 30\,fps. 
Since IMHD$^{2}$ and BEHAVE contain many long ($\geq 60$\,s) sequences whereas OMOMO clips are about 5\,s, we randomly crop IMHD$^{2}$ and BEHAVE into 5–10\,s windows during training and evaluation, which balances sequence length and increases sample diversity. 
Unless otherwise stated, we adopt an 80/20 train/test split at the sequence level; for IMHD$^{2}$ and BEHAVE, cropping is performed \emph{after} the split to avoid leakage. 

\begin{figure}[t]
  \centering
  \includegraphics[width=0.48\linewidth]{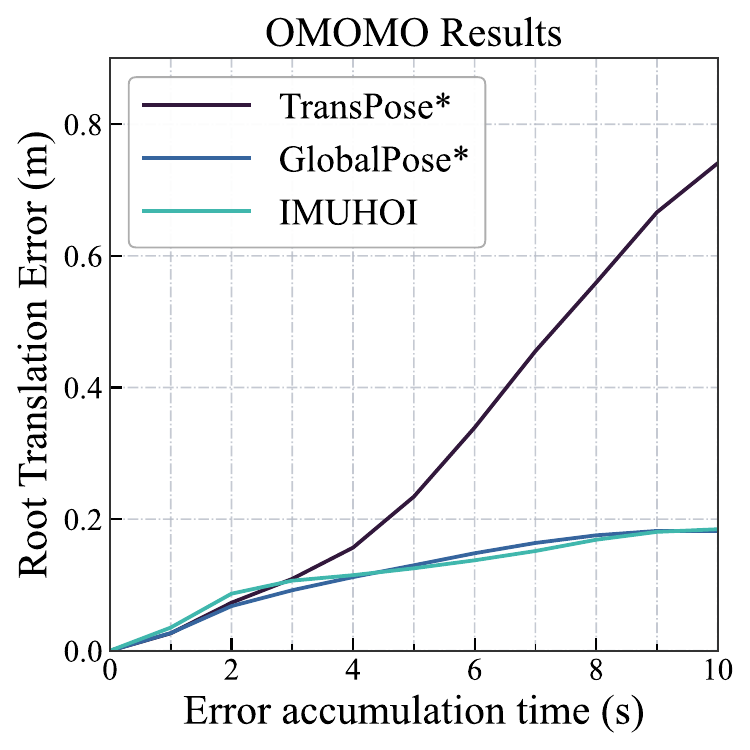}%
  \hfill
  \includegraphics[width=0.48\linewidth]{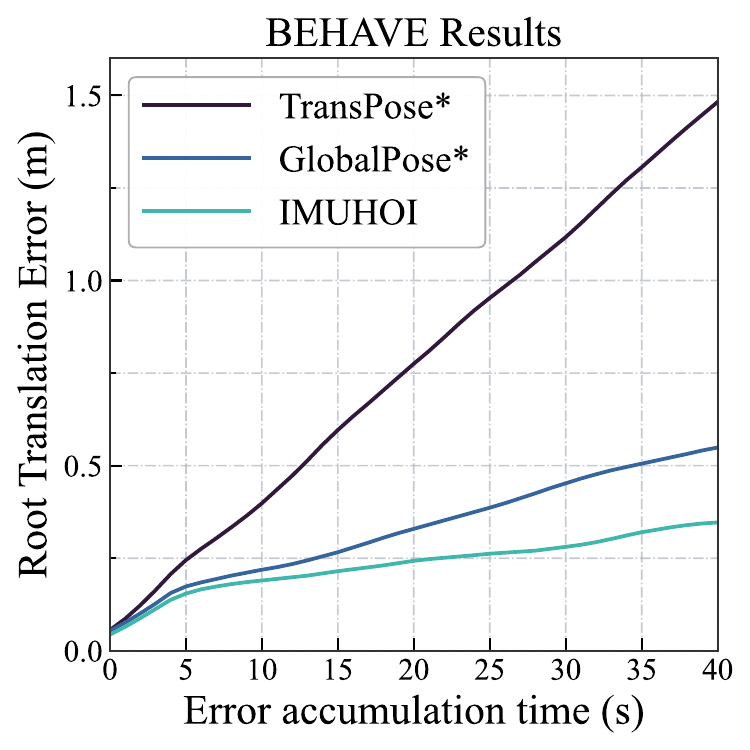}%
  \vspace{-3mm}
  \caption{Cumulative root-translation error vs.\ time for OMOMO (left) and BEHAVE (right).}
  \vspace{-3mm}
  \label{fig:curve_trans_error}
\end{figure}

\begin{figure*}[t]
    \centering
    \includegraphics[width=1.0\linewidth]{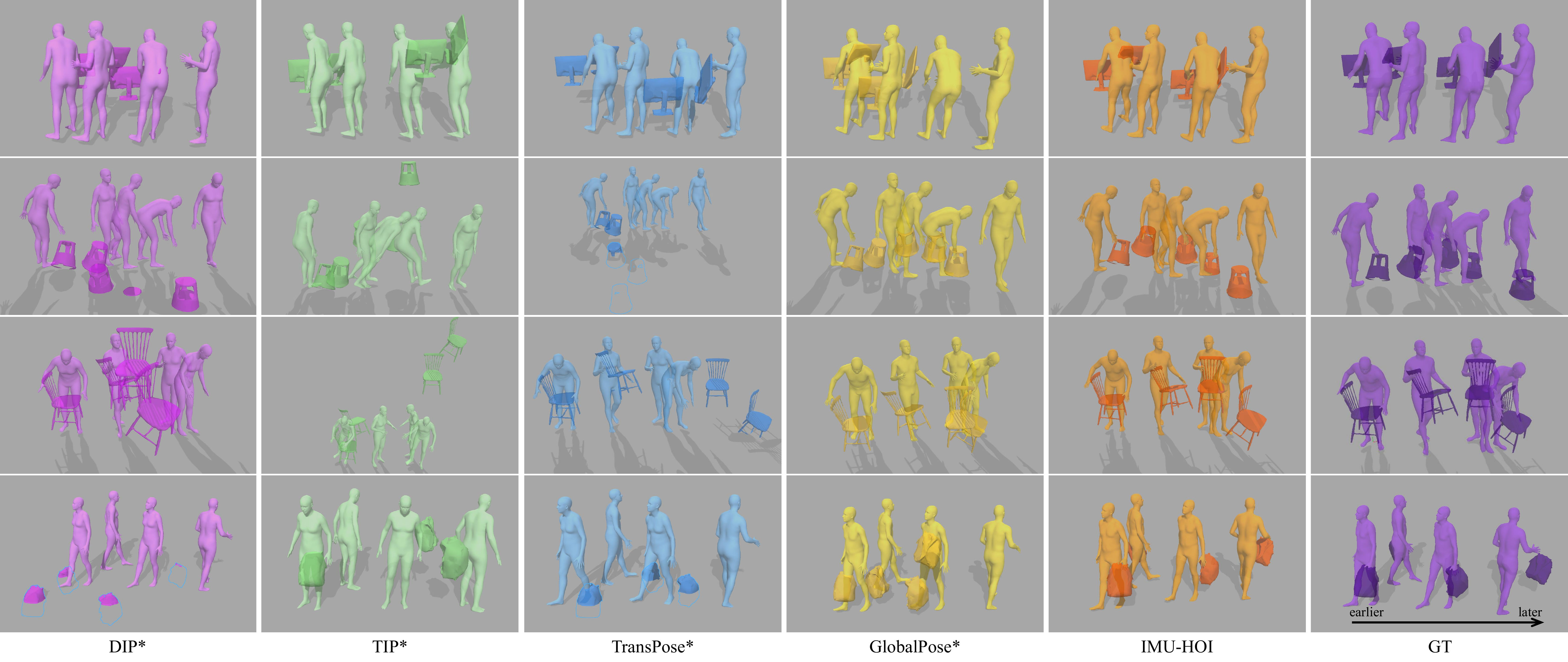}
    \vspace{-5mm}
    \caption{Qualitative comparison of motion estimation on four sequences from the BEHAVE test set.}
    \label{fig:qualitative_compare}
    \vspace{-5mm}
\end{figure*}


\medskip\noindent\textbf{Competing methods.}
We compare against four IMU-based motion capture baselines:
DIP~\cite{huang2018deep},
TIP~\cite{jiang2022transformer},
TransPose~\cite{yi2021transpose},
and GlobalPose~\cite{globalpose_2025}.
For a fair HOI comparison, we also report ``*'' variants (e.g., DIP*), where each backbone is augmented with an object-IMU branch and an object-velocity head trained under our HOI losses.
Each method is adapted to the HOI setting using its official backbone and training recipe, with our HOI-specific heads and losses applied in a consistent way across all baselines. 



\medskip\noindent\textbf{Metrics.}
Following prior work~\cite{yi2021transpose,petrov2023object}, we report six metrics:
\emph{Ang Err} [$^\circ$] and \emph{Pos Err} [cm] measure joint rotation and position accuracy;
\emph{Jitter} [mm/s$^3$] quantifies motion smoothness via mean joint jerk;
\emph{Trans Err} [cm] evaluates root translation accuracy;
\emph{Obj Err} [cm] measures object trajectory error;
and \emph{HOI Err} [cm] captures hand–object interaction accuracy on contact frames. 

Detailed information about datasets, competing methods and metrics can be found in the supplementary material.

\begin{table}[t]
\centering
\caption{Ablation of our object-translation heads. Lower is better. Best in \textbf{bold}.}
\vspace{-3mm}
\label{tab:ablation_heads}
\setlength{\tabcolsep}{4pt}
\renewcommand{\arraystretch}{1.05}
\scriptsize
\resizebox{\columnwidth}{!}{%
\begin{tabular}{lcccccc}
\toprule
& \multicolumn{2}{c}{OMOMO}
& \multicolumn{2}{c}{IMHD$^{2}$}
& \multicolumn{2}{c}{BEHAVE} \\
\cmidrule(lr){2-3}\cmidrule(lr){4-5}\cmidrule(lr){6-7}
\textbf{Methods} 
& Obj Err & HOI Err
& Obj Err & HOI Err
& Obj Err & HOI Err \\
\midrule
Our Method -- FK
& 31.06 & 32.51
& 61.58 & 57.64
& 23.68 & \textbf{24.58} \\
Our Method -- IMU Only
& 11.52 & 18.07
& 68.59 & 72.58
& 22.99 & 28.03 \\
\textbf{Our Method -- Fusion}
& \textbf{11.31} & \textbf{17.44}
& \textbf{43.97} & \textbf{43.39}
& \textbf{20.90} & 25.74 \\
\bottomrule
\end{tabular}%
}
\vspace{-1.5em}
\end{table}

\subsection{Comparison with State-of-the-Art}

We categorize prior methods~\cite{yi2021transpose, zhang2024dynamic} into two training–evaluation protocols, reflecting that most contemporary approaches do not estimate global/root translation:
(i) \emph{w/o root translation} (Tab.~\ref{tab:sota_without_root}); and
(ii) \emph{with root translation} (Tab.~\ref{tab:sota_with_root}).
Under protocol (i), for methods that originally include a translation head, we ablate that component and substitute the root translation with ground truth (GT) to isolate pose quality; the resulting models are re-trained and evaluated on the target datasets for fair comparison.
Under protocol (ii), translation-capable baselines are kept intact and trained/evaluated end-to-end.
This setup yields apples-to-apples comparisons across methods, regardless of whether they explicitly model root translation.

\medskip\noindent\textbf{Quantitative analysis.}
We first consider the setting where root translation estimation is disabled.
Across all three datasets in Tab.~\ref{tab:sota_without_root}, our method attains the lowest Obj Err and HOI Err in nearly every case, e.g., on OMOMO ($14.15 / 14.94$ for Obj / HOI), IMHD$^{2}$ ($49.76 / 51.09$), and BEHAVE ($22.26 / 22.62$).
Compared to GlobalPose*, our method achieves substantial Obj Err reductions of $25.19$ ($64.0\%$) / $51.51$ ($50.9\%$) / $17.88$ ($44.5\%$), and HOI Err reductions of $24.57$ ($62.2\%$) / $50.94$ ($49.9\%$) / $17.59$ ($43.7\%$) on OMOMO / IMHD$^{2}$ / BEHAVE, respectively.
These two metrics directly reflect the fidelity of the estimated object trajectory and human–object relations; the consistent gains show that explicitly modeling contact and jointly reasoning over human- and object-IMUs is crucial even when root translation is not evaluated.
Beyond interaction-centric scores, our method also achieves state-of-the-art Ang Err and Pos Err across all datasets, e.g., Ang Err of $2.84 / 4.51 / 4.52$ and Pos Err of $2.27 / 3.85 / 2.72$ on OMOMO / IMHD$^{2}$ / BEHAVE.

When root translation is evaluated (Tab.~\ref{tab:sota_with_root}), our method maintains clear advantages on Obj Err and HOI Err, while also delivering competitive root-translation accuracy.
On OMOMO, our Trans Err is on par with GlobalPose*, within $0.27$ ($2.5\%$) of its score; on the smaller IMHD$^{2}$ set we reduce Trans Err by $39.99$ ($68.5\%$), and on BEHAVE by $4.91$ ($30.6\%$).
This contrast suggests that GlobalPose* benefits strongly from large-scale training, whereas our contact-gated, RNN-based estimator generalizes better under limited data.
Notably, under the \emph{with root translation} setting, we focus on position metrics, so Tab.~\ref{tab:sota_with_root} omits angular pose errors, even though our method still maintains an advantage in those metrics.

Fig.~\ref{fig:curve_trans_error} plots cumulative root-translation error over time.
On BEHAVE our curve stays consistently below both TransPose* and GlobalPose*, and on OMOMO we clearly outperform TransPose* and remain on par with GlobalPose*.
Beyond accuracy, our model runs an order of magnitude faster than GlobalPose*: the latter relies on per-frame physical optimization in its forward pass, whereas our network is a lightweight causal RNN that performs a single feed-forward pass per step, enabling practical deployment.

\medskip\noindent\textbf{Qualitative analysis.}
Fig.~\ref{fig:qualitative_compare} visualizes results on untrimmed sequences from the BEHAVE test set.
We keep the original sequence lengths to stress-test \emph{generalization} and \emph{long-horizon stability} — a regime in which body- and object-IMUs inevitably accumulate drift.
Each column corresponds to a method (DIP*, TIP*, TransPose*, GlobalPose*, IMU-HOI, and GT), and rows show different interaction sequences ordered from earlier to later in time (indicated by the arrow in the GT column).
Across the first three rows (typical daily interactions), the extended baselines progressively lose hand–object alignment:
the object gradually slides away from the hand, contact is intermittently broken, and the relative human–object placement becomes inconsistent with the scene.
In contrast, IMU-HOI preserves the grasp configuration and the hand–object distance over time, matching the ground-truth progression.

The fourth row is a particularly challenging long-horizon case.
Here, methods that rely purely on IMUs for object pose (all baselines) suffer pronounced \emph{global drift}: the object trajectory deviates substantially and the interaction collapses after a long run.
IMU-HOI remains stable for the entire sequence, keeping the object consistently attached to the interacting hand and maintaining a plausible human–object relative pose.
We attribute this robustness to the proposed \emph{contact-gated fusion}: when confident hand contact is detected, the FK branch anchors the object to the hand; when contact weakens, the IMU branch provides complementary motion, and the gate mediates smooth transitions.

Overall, the qualitative trends in Fig.~\ref{fig:qualitative_compare} echo our quantitative gains:
on long sequences where IMU drift is unavoidable, IMU-HOI suppresses drift and preserves interaction semantics, whereas the extended HPE baselines eventually lose contact or accumulate large trajectory errors.

\begin{table}[t]
\centering
\caption{Fusion on off-the-shelf IMU HPE backbones. Lower is better. Best in \textbf{bold}.}
\vspace{-3mm}
\label{tab:plug_backbone}
\setlength{\tabcolsep}{4pt}
\renewcommand{\arraystretch}{1.06}
\scriptsize
\resizebox{\columnwidth}{!}{%
\begin{tabular}{lcccccc}
\toprule
& \multicolumn{2}{c}{OMOMO}
& \multicolumn{2}{c}{IMHD$^{2}$}
& \multicolumn{2}{c}{BEHAVE} \\
\cmidrule(lr){2-3}\cmidrule(lr){4-5}\cmidrule(lr){6-7}
\textbf{Methods}
& Obj Err & HOI Err
& Obj Err & HOI Err
& Obj Err & HOI Err \\
\midrule
DIP*                      & 25.84 & 26.79 & 90.37 & 93.09 & 38.91 & 39.31 \\
DIP* \;--\; Fusion        & \textbf{13.95} & \textbf{15.59} & \textbf{46.11} & \textbf{50.64} & \textbf{22.41} & \textbf{24.96} \\
\midrule
TransPose*                & 32.54 & 32.73 & 90.37 & 90.61 & 40.39 & 40.40 \\
TransPose* \;--\; Fusion  & \textbf{14.03} & \textbf{29.52} & \textbf{47.16} & \textbf{52.55} & \textbf{22.44} & \textbf{35.07} \\
\midrule
GlobalPose*               & 39.34 & 39.51 & 101.27 & 102.03 & 40.14 & 40.21 \\
GlobalPose* \;--\; Fusion & \textbf{14.22} & \textbf{20.60} & \textbf{46.70} & \textbf{53.69} & \textbf{22.69} & \textbf{27.98} \\
\bottomrule
\end{tabular}%
}
\vspace{-5mm}
\end{table}

\medskip\noindent\textbf{Effectiveness of the object-translation heads.}
Tab.~\ref{tab:ablation_heads} compares three variants: \emph{FK only}, \emph{IMU only}, and our full \emph{Fusion}.
Fusion consistently yields the best Obj Err on all datasets and the best (or near-best) HOI Err on two out of three datasets.
Concretely, relative to the stronger single head on each dataset, Fusion reduces Obj Err by $0.21$ ($1.8\%$) and HOI Err by $0.63$ ($3.5\%$) on OMOMO (vs.\ IMU-only);
by a much larger margin on the smaller IMHD$^{2}$ set, Obj Err drops $17.61$ ($28.6\%$) and HOI Err drops $14.25$ ($24.7\%$) (vs.\ FK-only);
and on BEHAVE our Fusion improves Obj Err by $2.09$ ($9.1\%$) over IMU-only while achieving the second-best HOI Err, within $4.7\%$ of FK-only.
These gains indicate that neither IMU-only (accurate but drifting) nor FK-only (anchored but locally biased) suffices alone; combining them under contact gating is crucial for robust object estimation and interaction fidelity.

\begin{figure}[t]
    \centering
    \includegraphics[width=1.0\linewidth]{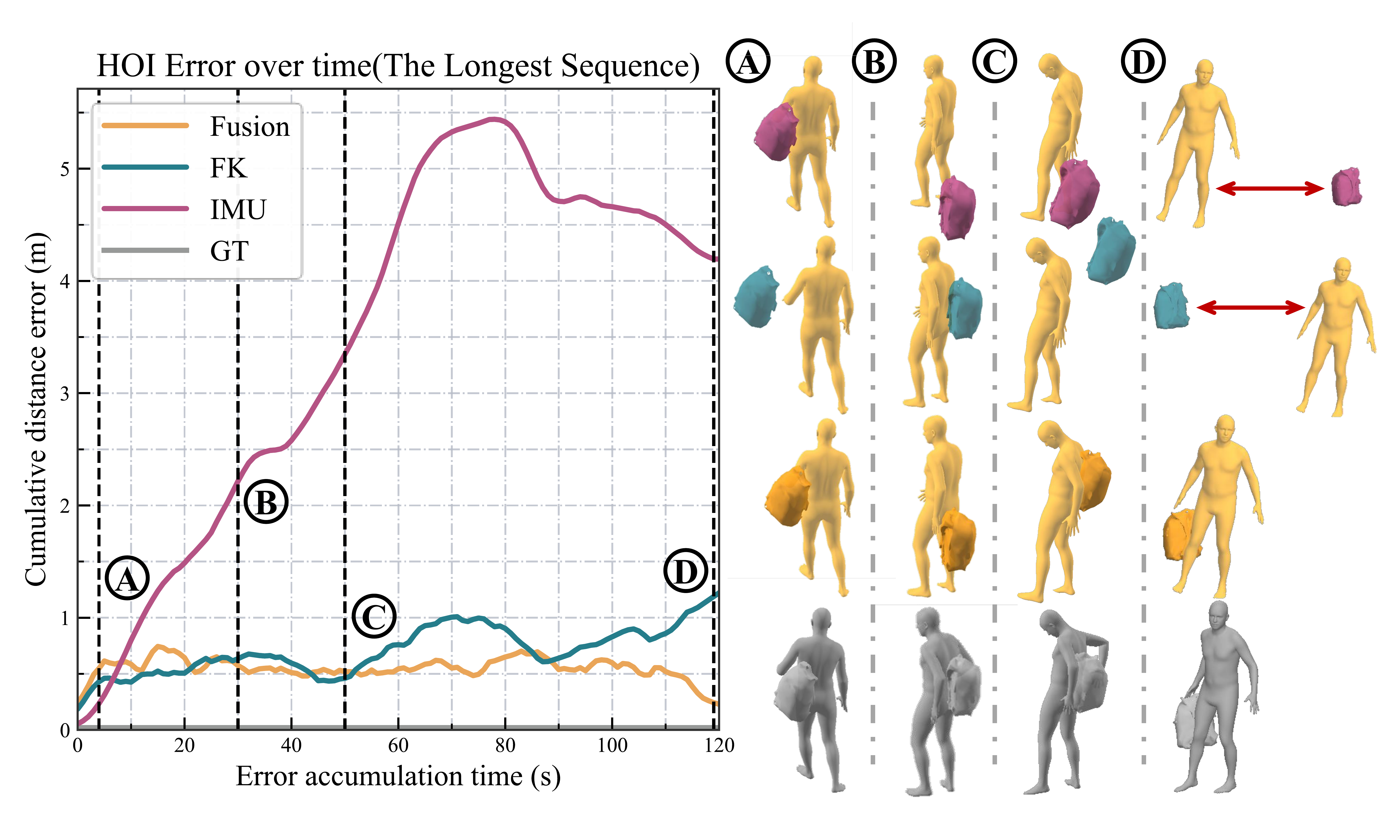}
    \vspace{-5mm}
    \caption{Visualization of error–time curves and reference frames (A–D) for the three object-translation heads.}
    \label{fig:curve_hoi_error}
    \vspace{-5mm}
\end{figure}

\medskip\noindent\textbf{Error–time curves and crossover behavior.}
To further probe the dynamics, we plot cumulative errors for the three heads and mark four reference times \textbf{A–D} (Fig.~\ref{fig:curve_hoi_error}); the right-hand panels show the corresponding visualizations (each with four people–object pairs).
At \textbf{A} (short horizon), the \emph{IMU-only} branch exhibits the lowest error—its raw inertial integration excels at capturing high-frequency, short-range motion with high instantaneous accuracy.
By \textbf{B} (longer horizon), \emph{FK-only} becomes preferable: anchoring the object to the hand through kinematics/contact prevents early drift, so its curve grows more slowly than IMU-only.
At \textbf{C}, our \emph{Fusion} overtakes both—contact-gated routing leverages FK when hand–object contact is confident (to avoid slippage) while still injecting IMU cues for short-term precision, producing a curve below both single heads.
Finally, by \textbf{D} (sequence end, $\approx 120$\,s), neither IMU-only nor FK-only preserves attachment—the object has clearly separated from the hand on both, which we attribute to accumulated errors caused by occasional contact-label inaccuracies that shift the interaction point and compound over time.
In contrast, \emph{Fusion} remains stable and attached by reweighting branches when contact confidence changes and by enforcing kinematic–inertial consistency, explaining the large long-horizon advantage observed in Tab.~\ref{tab:ablation_heads}.

\subsection{Ablation Studies}
\label{subsec:ablation}

\medskip\noindent\textbf{Backbone-agnostic plug-in study.}
Because the human module in our pipeline is replaceable, we compare simply augmenting each HPE backbone with an object head against plugging the same backbone into our contact-gated Fusion (Tab.~\ref{tab:plug_backbone}).
Across all datasets, the Fusion plug-in yields clear improvements on object-centric metrics over the dimension-augmented baselines.
In particular, for GlobalPose* on IMHD$^{2}$, Obj Err is reduced by $54.57$ ($53.9\%$) and HOI Err by $48.34$ ($47.4\%$), highlighting that our plug-and-play fusion substantially upgrades existing HPE models for HOI.

\section{Conclusion}
\label{sec:conclusion}

We presented \textbf{IMU-HOI}, a con
tact-conscious inertial fusion framework that jointly estimates full-body human pose and 6-DoF object trajectories from only sparse IMUs mounted on the body and a single manipulated object. By treating hand–object contact as a first-class probabilistic signal and routing between kinematic and inertial branches, our three-stage pipeline yields coherent, drift-resistant reconstructions of both human and object motion. Extensive experiments on three challenging HOI benchmarks demonstrate that IMU-HOI consistently improves object translation and interaction fidelity (Obj Err and HOI Err) over strong IMU-based baselines, while also achieving state-of-the-art human pose accuracy. Moreover, our design is modular: the human module can be instantiated by off-the-shelf HPE backbones (e.g., DIP, TransPose, GlobalPose), and our contact-gated fusion can be plugged in to equip them with robust object tracking capabilities.
\textbf{Limitations. }Our interaction model relies on a quasi-rigid, single-contact abstraction and does not explicitly handle complex phenomena such as sliding contacts, multiple simultaneous contact points, or interactions with deformable objects.

\newpage
{
    \small
    \bibliographystyle{ieeenat_fullname}
    \bibliography{ref}
}


\end{document}